\documentclass[11pt,a4paper]{article}
\usepackage{graphicx}
\usepackage[hyperref]{acl2020}
\usepackage{times}
\usepackage{latexsym}
\usepackage{amsmath}
\usepackage{lipsum}
\usepackage{mathtools}
\usepackage{cuted}
\usepackage{pifont}
\newcommand{\cmark}{\ding{51}}%
\newcommand{\xmark}{\ding{55}}%

\usepackage{microtype}

\aclfinalcopy 

\makeatletter
\newcommand\footnoteref[1]{\protected@xdef\@thefnmark{\ref{#1}}\@footnotemark}
\makeatother
\newcommand{\printfnsymbol}[1]{%
  \textsuperscript{\@fnsymbol{#1}}%
}
\title{DIET: Lightweight Language Understanding for Dialogue Systems}
\author{Tanja Bunk\textsuperscript{1}\thanks{\hspace*{0.5em}t.bunk@rasa.com} \qquad Daksh Varshneya\textsuperscript{1}\thanks{\hspace*{0.5em}d.varshneya@rasa.com} \qquad Vladimir Vlasov\textsuperscript{1}\thanks{\hspace*{0.5em}vladimir@rasa.com} \qquad Alan Nichol\thanks{\hspace*{0.5em}alan@rasa.com} \\ Rasa}

\date{}

\begin{document}
\maketitle
\begin{abstract}
Large-scale pre-trained language models have shown impressive results on language understanding benchmarks like GLUE and SuperGLUE, improving considerably over other pre-training methods like distributed representations (GloVe) and purely supervised approaches. We introduce the Dual Intent and Entity Transformer (DIET) architecture, and study the effectiveness of different pre-trained representations on intent and entity prediction, two common dialogue language understanding tasks. DIET advances the state of the art on a complex multi-domain NLU dataset and achieves similarly high performance on other simpler datasets. Surprisingly, we show that there is no clear benefit to using large pre-trained models for this task, and in fact DIET improves upon the current state of the art even in a purely supervised setup without any pre-trained embeddings. Our best performing model outperforms fine-tuning BERT and is about six times faster to train. 
\end{abstract}

\section{Introduction}
\label{sec:introduction}

    \footnotetext[1]{authors have equally contributed}
    Two common approaches to data-driven dialogue modeling are the end-to-end and the modular systems.
    Modular approaches like POMDP-based dialogue policies~\citep{williams2007partially} and Hybrid Code Networks~\citep{williams2017hybrid}
    use separate natural language understanding (NLU) and generation (NLG) systems. The dialogue policy itself receives the output from the NLU system and chooses the next system action, before the NLG system generates a corresponding response. 
    In the end-to-end approach user input is directly fed into the dialogue policy to predict the next system utterance.
    Recently these two approaches have been combined in Fusion Networks~\citep{mehri2019structured}.
    
    In the context of dialogue systems, natural language understanding typically refers to two subtasks: intent classification and entity recognition. 
    
    ~\citeauthor{goo2018slot} argue that modeling these sub-tasks separately can suffer from error propagation and hence a single multi-task architecture should benefit from mutual enhancement between two tasks.

    Recent work has shown that  large pre-trained language models yield the best performance on challenging language understanding benchmarks (see section \ref{sec:related_work}). However, the computational cost of both pre-training and fine-tuning such models is considerable~\citep{strubell2019energy}.

    Dialogue systems are not only developed by researchers, but by many thousands of software developers worldwide. Facebook's Messenger platform alone supports hundreds of thousands of third party conversational assistants~\cite{khari}.
    For these applications it is desirable that models can be trained and iterated upon quickly to fit into a typical software development workflow. 
    Furthermore, since many of these assistants operate in languages other than English, it is important to understand what performance can be achieved \emph{without} large-scale pre-training.
    
    In this paper, we propose DIET (Dual Intent and Entity Transformer), a new multi-task architecture for intent classification and entity recognition. One key feature is the ability to incorporate pre-trained word embeddings from language models and combine these with sparse word and character level n-gram features in a plug-and-play fashion. Our experiments demonstrate that even without pre-trained embeddings, using only sparse word and character level n-gram features, DIET improves upon the current state of the art on a complex NLU dataset. Moreover, adding pre-trained word and sentence embeddings from language models further improves the overall accuracy on all tasks. Our best performing model significantly outperforms fine-tuning BERT and is six times faster to train. 
    Documented code to reproduce these experiments is available online at \url{https://github.com/RasaHQ/DIET-paper}.

\section{Related Work}
\label{sec:related_work}

\subsection{Transfer learning of dense representations}
Top performing models~\citep{mtdnn2019,ernie2019} on language understanding benchmarks like GLUE~\citep{glue} and SuperGLUE~\citep{superglue2019} benefit from using dense representations of words and sentences from large pre-trained language models like ELMo~\citep{elmo}, BERT~\citep{devlin2018bert}, GPT~\citep{gpt} etc. Since these embeddings are trained on large scale natural language text corpora, they generalize well across tasks and can be transferred as input features to other language understanding tasks with or without fine-tuning~\citep{elmo,videobert,patentbert,docbert,commonsensebert}. Different fine-tuning strategies have also been proposed for effective transfer learning across tasks~\citep{howardruder,howtotunebert}. However, \citet{ruder2019tuning} show that fine-tuning a large pre-trained language model like BERT may not be optimal for every downstream task. Moreover, these large scale language models are slow, expensive to train and hence not ideal for real-world conversational AI applications~\citep{henderson2019convert}. To achieve a more compact model, \citet{henderson2019convert} pre-train a word and sentence level encoder on a large scale conversational corpus from Reddit~\citep{redditdata}. The resultant sentence level dense representations, when transferred (without fine-tuning) to a downstream task of intent classification, perform much better than embeddings from BERT and ELMo. 
We further investigate this behaviour for the task of joint intent classification and entity recognition. We also study the impact of using sparse representations like word level one-hot encodings and character level n-grams along with dense representations transferred from large pre-trained language models.

\subsection{Joint Intent Classification and Named Entity Recognition}
In recent years a number of approaches have been studied for training intent classification and named entity recognition (NER) in a multi-task setup. \citet{zhang2016ajoint} proposed a joint architecture composed of a Bidirectional Gated Recurrent Unit (BiGRU). The hidden state of each time step is used for entity tagging and the hidden state of last time step is used for intent classification. \citet{liu2016attention,jointvarghese} and~\citet{goo2018slot} propose an attention-based Bidirectional Long Short Term Memory (BiLSTM) for joint intent classification and NER. \citet{haihong2019anovel} introduce a  co-attention network on top of individual intent and entity attention units for mutual information sharing between each task. \citet{Chen2019} propose Joint BERT which is built on top of BERT and is trained in an end to end fashion. They use the hidden state of the first special token \verb+[CLS]+ for intent classification. The entity labels are predicted using the final hidden states of other tokens. A hierarchical bottom-up architecture was proposed by~\citet{vanzo2019hermit} composed of BiLSTM units to capture shallower representations of semantic frames~\citep{baker97}. They predict dialogue acts, intents and entity labels from representations learnt by individual layers stacked in a bottom-up fashion. In this work, we adopt a similar transformer-based multi-task setup for DIET and also perform an ablation study to observe its effectiveness compared to a single task setup.

\section{DIET Architecture}
\label{sec:architecture}

    \begin{figure*}
    \centering
        \includegraphics[width=0.9\linewidth]{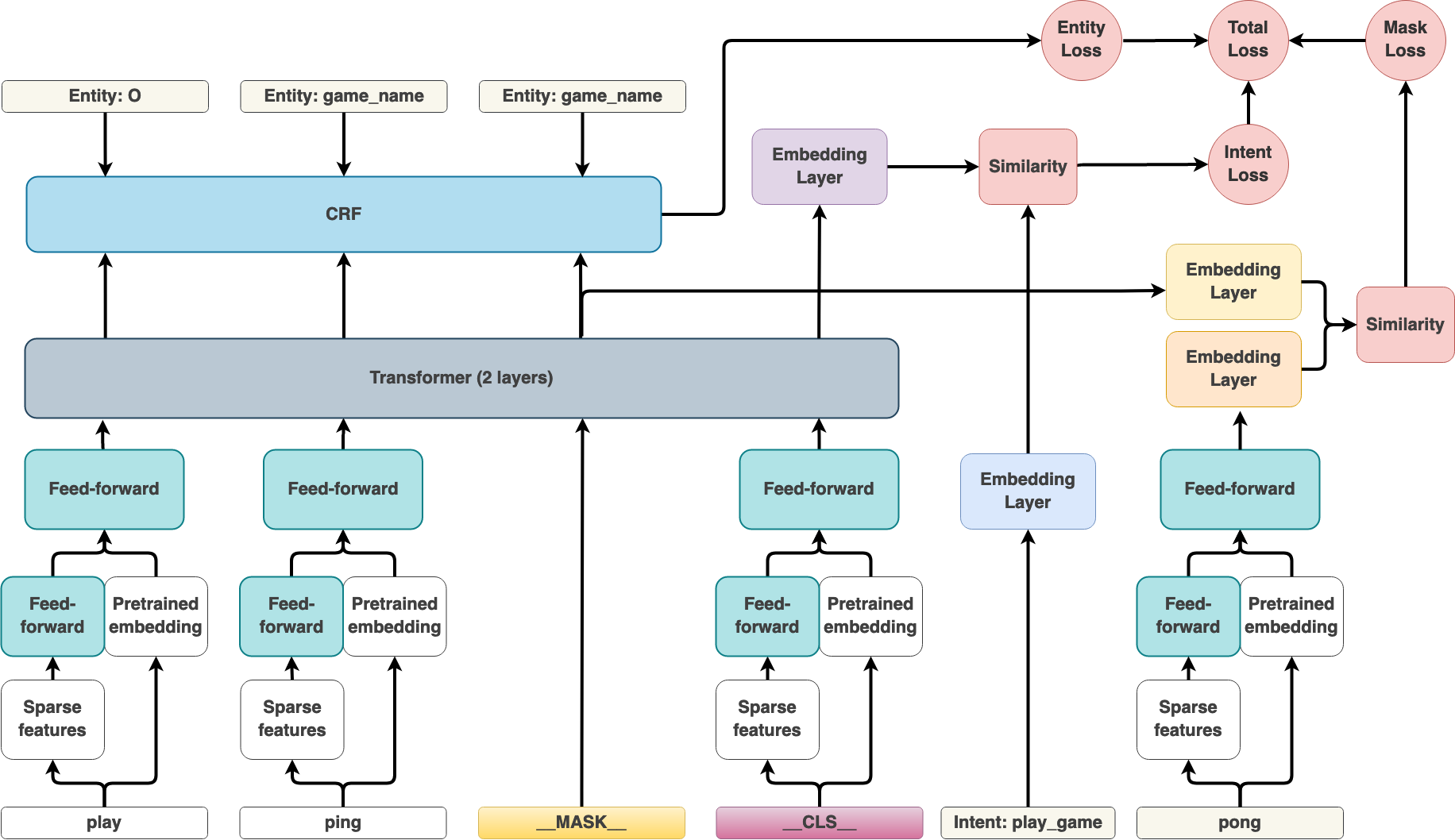}
        \caption{A schematic representation of the DIET architecture. The phrase "play ping pong" has the intent \texttt{play\_game} and entity \texttt{game\_name} with value "ping pong". Weights of the feed-forward layers are shared across tokens. }
        \label{fig:architecture}
    \end{figure*}

    A schematic representation of our architecture is illustrated in Figure~\ref{fig:architecture}. DIET consists of several key parts.
    
    \paragraph{Featurization}
    Input sentences are treated as a sequence of tokens, which can be either words or subwords depending on the featurization pipeline. Following~\citet{devlin2018bert}, we add a special classification token \verb+__CLS__+ to the end of each sentence. Each input token is featurized with what we call sparse features and/or dense features. Sparse features are token level one-hot encodings and multi-hot encodings of character  n-grams $(n\leq 5)$. Character n-grams contain a lot of redundant information, so to avoid overfitting we apply dropout to these sparse features. Dense features can be any pre-trained word embeddings: ConveRT~\citep{henderson2019convert}, BERT~\citep{devlin2018bert} or GloVe~\citep{pennington2014glove}. Since ConveRT is also trained as a sentence encoder, when using ConveRT we set the initial embedding for \verb+__CLS__+ token as the sentence encoding of the input sentence obtained from ConveRT.\footnote{Sentence embeddings from ConveRT are 1024-dimensional and word embeddings are 512-dimensional. To overcome this dimension mismatch, we use a simple trick of tiling the word embeddings to extra 512 dimensions and get 1024-dimensional word embeddings. This keeps the neural architecture the same for different pre-trained embeddings.} This adds extra contextual information for the complete sentence in addition to information from individual word embeddings. For out-of-the-box pre-trained BERT, we set it to the corresponding output embedding of the BERT \verb+[CLS]+ token and for GloVe, to the mean of the embeddings of the tokens in a sentence.
    Sparse features are passed through a fully connected layer with shared weights across all sequence steps to match the dimension of the dense features. The output of the fully connected layer is concatenated with the dense features from pre-trained models.

    \paragraph{Transformer}
    To encode context across the complete sentence, we use a 2 layer transformer~\citep{vaswani2017attention} with relative position attention~\citep{shaw2018self}. The transformer architecture requires its input to be the same dimension as the transformer layers. Therefore, the concatenated features are passed through another fully connected layer with shared weights across all sequence steps to match the dimension of the transformer layers, which in our experiments is $256$.
    
    \paragraph{Named entity recognition}
    A sequence of entity labels~$\pmb{y}_{\text{entity}}$ is predicted through a Conditional Random Field (CRF)~\citep{lafferty2001conditional} tagging layer on top of the transformer output sequence $\pmb{a}$ corresponding to an input sequence of tokens.
    \begin{equation}
        L_{\text{E}} = L_\text{CRF}(\pmb{a}, \pmb{y}_{\text{entity}}),
        \label{eq:entity}
    \end{equation}
    where $L_\text{CRF}(.)$ denotes negative log-likelihood for a CRF~\citep{crfloss}.

    \paragraph{Intent classification}
    The transformer output for \verb+__CLS__+ token $a_{\text{CLS}}$ and intent labels $y_{\text{intent}}$ are embedded into a single semantic vector space $h_{\text{CLS}} = E(a_{\text{CLS}})$,  $h_{\text{intent}} = E(y_{\text{intent}})$, where $h\in{\rm I\!R}^{20}$. We use the dot-product loss~\citep{wu2017starspace, henderson2019training, vlasov2019dialogue} to maximize the similarity $S_{\text{I}}^+ = h_{\text{CLS}}^T h_{\text{intent}}^+$ with the target label $y_{\text{intent}}^+$ and minimize similarities $S_{\text{I}}^- = h_{\text{CLS}}^T h_{\text{intent}}^-$ with negative samples $y_{\text{intent}}^-$.
    \begin{equation}
        L_{\text{I}} = - \biggl\langle S_{\text{I}}^+ - \log\biggl(e^{S_{\text{I}}^+} + \sum_{\Omega_{\text{I}}^-}e^{S_{\text{I}}^-}\biggr) \biggr\rangle,
        \label{eq:intent}
    \end{equation}
    where the sum is taken over the set of negative samples $\Omega_{\text{I}}^-$ and the average $\langle . \rangle$ is taken over all examples.
    
    At inference time, the dot-product similarity serves as a ranker over all possible intent labels.
    
    \paragraph{Masking}
    Inspired by the masked language modelling task~\citep{taylor1953cloze, devlin2018bert}, we add an additional training objective to predict randomly masked input tokens. We select at random $15\%$ of the input tokens in a sequence. For a selected token, in $70\%$ of cases we substitute the input with the vector corresponding to the special mask token \verb+__MASK__+, in $10\%$ of  cases we substitute the input with the vector corresponding to a random token and in the remaining $20\%$ we keep the original input. The output of the transformer $a_{\text{MASK}}$ for each selected token $y_{\text{token}}$ is fed through a dot-product loss~\citep{wu2017starspace, henderson2019training, vlasov2019dialogue}  similar to the intent loss.
    \begin{equation}
        L_{\text{M}} = - \biggl\langle S_{\text{M}}^+ - \log\biggl(e^{S_{\text{M}}^+} + \sum_{\Omega_{\text{M}}^-}e^{S_{\text{M}}^-}\biggr) \biggr\rangle,
        \label{eq:mask}
    \end{equation}
    where $S_{\text{M}}^+ = h_{\text{MASK}}^T h_{\text{token}}^+$ is the similarity with the target label $y_{\text{token}}^+$ and $S_{\text{M}}^- = h_{\text{MASK}}^T h_{\text{token}}^-$ are the similarities with negative samples $y_{\text{token}}^-$, $h_{\text{MASK}} = E(a_{\text{MASK}})$ and $h_{\text{token}} = E(y_{\text{token}})$ are the corresponding embedding vectors $h\in{\rm I\!R}^{20}$; the sum is taken over the set of negative samples $\Omega_{\text{M}}^-$ and the average $\langle . \rangle$ is taken over all examples.

    We hypothesize that adding a training objective for reconstructing masked input should act as a regularizer as well as help the model learn more general features from text and not only discriminative features for classification~\citep{class-reconstruct2019}.

    \paragraph{Total loss}
    We train the model in multi-task fashion by minimizing the total loss $L_{total}$.
    \begin{equation}
        L_{total} = L_{I} + L_{E} + L_{M}
        \label{eq:total_loss}
    \end{equation}
    The architecture can be configured to turn off any of the losses in the sum above.

    \paragraph{Batching}
    We use a balanced batching strategy~\citep{vlasov2019dialogue} to mitigate class imbalance~\citep{japkowicz2002class} as some intents can be more frequent than others.
    We also increase our batch size throughout training as another source of regularization~\citep{smith2017don}.

\section{Experimental Evaluation}
\label{sec:experimental_evaluation}

In this section we first describe the datasets used in our experiments, then we describe the experimental setup, followed by an ablation study to understand the effectiveness of each component of the architecture.

\subsection{Datasets}
\label{subsec:datasets}

We used three datasets for our evaluation: NLU-Benchmark, ATIS, and SNIPS. The focus of our experiments is the NLU-Benchmark dataset, since it is the most challenging of the three. The state of the art on ATIS and SNIPS is already close to 100\% test set accuracy, see Table~\ref{tab:results-snips-atis}.

    \paragraph{NLU-Benchmark dataset} The NLU-Benchmark dataset~\citep{liu2019benchmarking}, available online\footnote{\url{https://github.com/xliuhw/NLU-Evaluation-Data/}}, is annotated with scenarios, actions, and entities. For example, ``schedule a call with Lisa on Monday morning" is annotated with the scenario \texttt{calendar}, the action \texttt{set\_event}, and the entities [\texttt{event\_name}: \textit{a call with Lisa}] and [\texttt{date}: \textit{Monday morning}]. The intent label is obtained by concatenating the scenario and action labels (e.g. \texttt{calendar\_set\_event}). The dataset has 25,716 utterances which cover multiple home assistant tasks, such as playing music or calendar queries, chit-chat, and commands issued to a robot. The data is split into 10 folds. Each fold has its own train and test set of respectively 9960 and 1076 utterances.\footnote{Some utterances appear in multiple folds.} Overall 64 intents and 54 entity types are present.
    
    \paragraph{ATIS} ATIS~\citep{hemphill1990atis} is a well-studied dataset in the field of NLU. It comprises annotated transcripts of audio recordings of people making flight reservations. We used the same data split as~\citet{Chen2019}, originally proposed by~\citet{goo2018slot} and available online\footnote{\label{github-atis-snips}\url{https://github.com/MiuLab/SlotGated-SLU}}. The training, development, and test sets contain 4,478, 500 and 893 utterances. The training dataset has 21 intents and 79 entities.
    
    \paragraph{SNIPS} This dataset is collected from the Snips personal voice assistant~\citep{coucke2018snips}. It contains 13,784 training and 700 test examples. For fair comparison, we used the same data split as~\citet{Chen2019} and~\citet{goo2018slot}. 700 examples from the training set are used as development set. The data can be found online\footnoteref{github-atis-snips}. The SNIPS dataset contains 7 intents and 39 entities.
    

\subsection{Experimental Setup}
\label{subsec:experiment_setup}

Our model is implemented in Tensorflow~\citep{abadi2015tensorflow}. We used the first fold of the NLU-Benchmark dataset to select hyperparameters. We randomly took 250 utterances from the training set as a development set for that purpose. We trained our models over 200 epochs on a machine with 4 CPUs, 15 GB of memory and one NVIDIA Tesla K80. We used Adam~\citep{Adam2014} for optimization with an initial learning rate of 0.001. The batch size increased incrementally from 64 to 128~\citep{smith2017don}. Training our model on the first fold of the NLU-Benchmark dataset takes around one hour. At inference time we need around 80ms to process one utterance. 

\subsection{Experiments on NLU-Benchmark dataset} 
\label{subsec:experiment_nlu_benchmark}
The NLU-Benchmark dataset contains 10 folds, each with a separate train and test set. To obtain the overall performance of our model on this dataset we followed the approach of~\citet{vanzo2019hermit}: train 10 models independently, one for each fold and take the average as the final score. Micro-averaged precision, recall and F1 score are used as metrics. True positives, false positives, and false negatives for intent labels are calculated as in any other multi-class classification task. An entity counts as true positive if there is an overlap between the predicted and the gold span and their labels match.

\begin{table}
\centering
\begin{tabular}{llrr}
 & \multicolumn{1}{c}{} & \multicolumn{1}{c}{\textbf{Intent}} & \multicolumn{1}{c}{\textbf{Entities}} \\ 
\multicolumn{1}{l|}{}                  & F1 & 87.55$\pm$0.63          & 84.74$\pm$1.18 \\
\multicolumn{1}{l|}{HERMIT}            & R  & 87.70$\pm$0.64          & 82.04$\pm$2.12 \\
\multicolumn{1}{l|}{}                  & P  & 87.41$\pm$0.63          & \textbf{87.65$\pm$0.98} \\ \hline
\multicolumn{1}{l|}{sparse +}          & F1 & \textbf{90.18$\pm$0.53} & \textbf{86.04$\pm$1.01} \\
\multicolumn{1}{l|}{$\mathrm{ConveRT}^{\dagger}$} & R  & \textbf{90.18$\pm$0.53} & \textbf{86.13$\pm$0.99} \\
\multicolumn{1}{l|}{}                  & P  & \textbf{90.18$\pm$0.53} & 85.95$\pm$1.42
\end{tabular}
\caption{Results from HERMIT~\citep{vanzo2019hermit} and from our best performing configuration of DIET on the NLU-Benchmark dataset. Our best performing model uses word and character level sparse features and combines them with embeddings from ConveRT. The model does not use a mask loss (indicated by the $\dagger$).}\label{tab:results-nlu-benchmark}
\end{table}

Table~\ref{tab:results-nlu-benchmark} shows the results of our best performing model on the NLU-Benchmark dataset. 
Our best performing model uses sparse features, i.e. one-hot encodings at the token level and multi-hot encodings of character n-grams ($n\leq 5$). These sparse features are combined with dense embeddings from ConveRT~\citep{henderson2019convert}. 
Our best performing model does not use a mask loss (described in Section~\ref{sec:architecture} and indicated by $\dagger$ in the table). 
We outperform HERMIT on intents by over 2\% absolute.
Our micro-averaged F1 score on entities (86.04\%) is also higher than HERMIT (84.74\%). 
HERMIT reports a similar precision value on entities, however, our recall value is much higher (86.13\% compared to 82.04\%).

\subsection{Ablation Study on NLU-Benchmark dataset}
\label{subsec:ablation_study}

We used the NLU-Benchmark dataset to evaluate different components of our model architecture as it covers multiple domains and has the most number of intents and entities of the three datasets.

\begin{table}
\centering
\begin{tabular}{llrr}
\multicolumn{1}{c}{} & \multicolumn{1}{c}{} & \multicolumn{1}{c}{\textbf{Intent}} & \multicolumn{1}{c}{\textbf{Entities}} \\
\multicolumn{1}{l|}{single-task:}   & F1 & 90.90$\pm$0.19 & - \\
\multicolumn{1}{l|}{intent}         & R  & 90.90$\pm$0.19 & - \\
\multicolumn{1}{l|}{classification} & P  & 90.90$\pm$0.19 & - \\ \hline
\multicolumn{1}{l|}{single-task:}   & F1 & -              & 82.57$\pm$1.41 \\
\multicolumn{1}{l|}{entity}         & R  & -              & 81.85$\pm$1.87 \\
\multicolumn{1}{l|}{recognition}    & P  & -              & 83.32$\pm$1.51
\end{tabular}
\caption{Training DIET on just a single task, i.e. intent classification or entity recognition, on the NLU-Benchmark dataset. 
}\label{tab:results-single-task}
\end{table}

\paragraph{Importance of joint training} In order to evaluate if the two tasks, i.e. intent classification and named entity recognition, benefit from being optimized jointly or not, we trained models for each of the tasks individually. Table~\ref{tab:results-single-task} lists the results of just training a single task with DIET. 
The results show that the performance of intent classification slightly decreases when trained jointly with entity recognition (90.90\% vs 90.18\%). It should be noted that the best performing configuration for single task training for intent classification corresponds to using embeddings from ConveRT with no transformer layers\footnote{This result is in line with the results reported in \citet{casanueva2020efficient}}.
However, the micro-averaged F1 score of entities drops from 86.04\% to 82.57\% when entities are trained separately. 
Inspecting the NLU-Benchmark dataset, this is likely due to strong correlation between particular intents and the presence of specific entities. For example, almost every utterance that belongs to the \texttt{play\_game} intent has an entity called \texttt{game\_name}.
Also, the entity \texttt{game\_name} only occurs together with the intent \texttt{play\_game}. We believe that this result further brings out the importance of having a modular and configurable architecture like DIET in order to handle trade-off in performance across both tasks.

\begin{table*}
\centering
\begin{tabular}{c|c|c|rr}
sparse & dense & mask loss & \multicolumn{1}{c}{\textbf{Intent}} & \multicolumn{1}{c}{\textbf{Entities}} \\ \hline
\cmark     & \xmark  & \xmark     & 87.10$\pm$0.75 & 83.88$\pm$0.98 \\
\cmark     & \xmark  & \cmark     & 88.19$\pm$0.84 & 85.12$\pm$0.85 \\
\xmark     & GloVe   & \xmark     & 89.20$\pm$0.90 & 84.34$\pm$1.03 \\
\cmark     & GloVe   & \xmark     & 89.38$\pm$0.71 & 84.89$\pm$0.91 \\
\xmark     & GloVe   & \cmark     & 88.78$\pm$0.70 & 85.06$\pm$0.84  \\
\cmark     & GloVe   & \cmark     & 89.13$\pm$0.77 & 86.04$\pm$1.09 \\
\xmark     & BERT    & \xmark     & 87.44$\pm$0.92 & 84.20$\pm$0.91 \\
\cmark     & BERT    & \xmark     & 88.46$\pm$0.88 & 85.26$\pm$1.01 \\
\xmark     & BERT    & \cmark     & 86.92$\pm$1.09 & 83.96$\pm$1.33  \\
\cmark     & BERT    & \cmark     & 87.45$\pm$0.67 & 84.64$\pm$1.31 \\ 
\xmark     & ConveRT & \xmark     & 89.76$\pm$0.98 & \textbf{86.06$\pm$1.38} \\
\cmark     & ConveRT & \xmark     & \textbf{90.18$\pm$0.53} & 86.04$\pm$1.01 \\
\xmark     & ConveRT & \cmark     & 90.15$\pm$0.68 & 85.76$\pm$0.80  \\
\cmark     & ConveRT & \cmark     & 89.47$\pm$0.74 & 86.04$\pm$1.29 \\
\end{tabular}
\caption{Comparison of different featurization and architecture components on NLU-Benchmark dataset. The three columns on the left indicate whether sparse features are used or not, what kind of dense features are used, if any, and whether the model was trained with a mask loss or not. The reported numbers are micro-averaged F1 scores.}\label{tab:results-featurization}
\end{table*}

\paragraph{Importance of different featurization components and masking} As described in Section~\ref{sec:architecture} embeddings from different pre-trained language models can be used as dense features. 
We trained multiple variants to study the effectiveness of each: only sparse features, i.e. one-hot encodings at the token level and multi-hot encodings of character n-grams ($n\leq 5$), and combinations of those together with ConveRT, BERT, or GloVe. 
Additionally, we trained each combination with and without the mask loss.
The results presented in Table~\ref{tab:results-featurization} show F1 scores for both intent classification and entity recognition and indicate multiple observations:
DIET performance is competitive when using sparse features together with the mask loss, without any pre-trained embeddings. Adding a mask loss improves performance by around 1\% absolute on both intents and entities. DIET with GloVe embeddings is also equally competitive and is further enhanced on both intents and entities when used in combination with sparse features and mask loss. Interestingly, using contextual BERT embeddings as dense features performs worse than GloVe. We hypothesize that this is because BERT is pre-trained primarily on prose and hence requires fine-tuning before being transferred to a dialogue task. The performance of DIET with ConveRT embeddings supports this, since ConveRT was trained specifically on conversational data. ConveRT embeddings with the addition of sparse features achieves the best F1 score on intent classification and it outperforms the state of the art on both intent classification and entity recognition by a considerable margin of around 3\% absolute. Adding a mask loss seems to slightly hurt the performance when used with BERT and ConveRT as dense features.

\begin{table}
\centering
\begin{tabular}{llrr}
\multicolumn{1}{c}{} & \multicolumn{1}{c}{} & \multicolumn{1}{c}{\textbf{Intent}} & \multicolumn{1}{c}{\textbf{Entities}} \\
\multicolumn{1}{l|}{Fine-tuned}        & F1 & 89.67$\pm$0.48          & 85.73$\pm$0.91 \\
\multicolumn{1}{l|}{BERT}              & R  & 89.67$\pm$0.48          & 84.71$\pm$1.28 \\
\multicolumn{1}{l|}{}                  & P  & 89.67$\pm$0.48          & \textbf{86.78$\pm$1.02} \\ \hline
\multicolumn{1}{l|}{sparse +}          & F1 & \textbf{90.18$\pm$0.53} & \textbf{86.04$\pm$1.01} \\
\multicolumn{1}{l|}{$\mathrm{ConveRT}^{\dagger}$} & R  & \textbf{90.18$\pm$0.53} & \textbf{86.13$\pm$0.99} \\
\multicolumn{1}{l|}{}                  & P  & \textbf{90.18$\pm$0.53} & 85.95$\pm$1.42
\end{tabular}
\caption{Comparison of best performing feature set for DIET against fine-tunable BERT inside DIET on the NLU-Benchmark dataset. The best performing feature set for DIET contains sparse features combined with embeddings from ConveRT (not fined-tuned) without a mask loss (indicated by the $\dagger$). Fine-tuning BERT with DIET takes 60 hours as compared to just 10 hours for DIET with sparse and ConveRT features.}\label{tab:results-BERT}
\end{table}

\paragraph{Comparison with fine-tuned BERT} Following~\citet{ruder2019tuning}, we evaluate the effectiveness of incorporating BERT inside the featurization pipeline of DIET and fine-tuning the entire model. Table~\ref{tab:results-BERT} shows DIET with frozen ConveRT embeddings as dense features and word, char level sparse features outperforms fine-tuned BERT on entity recognition while performing on par for intent classification. This result is especially important because fine-tuning BERT inside DIET on all 10 folds of NLU-Benchmark dataset takes 60 hours, compared to 10 hours for DIET with embeddings from ConveRT and sparse features.

\subsection{Experiments on ATIS and SNIPS}
\label{subsec:experiment_atis_snips}

\begin{table*}
\centering
\begin{tabular}{l|rr|rr}
 & \multicolumn{2}{c|}{\textbf{ATIS}} & \multicolumn{2}{c}{\textbf{SNIPS}} \\
 & \multicolumn{1}{c}{Intent} & \multicolumn{1}{c|}{Entities} & \multicolumn{1}{c}{Intent} & \multicolumn{1}{c}{Entities} \\ \hline
Joint BERT                     & \textbf{97.90} & \textbf{96.10} & \textbf{98.60} & \textbf{97.00} \\ \hline
sparse + $\mathrm{ConveRT}^{\dagger\ast}$ & 96.59          & 95.08          & 98.03          & 94.79 \\
sparse + $\mathrm{GloVe}^{\ast}$          & 96.31          & 94.99          & 97.50          & 94.84 \\
$\mathrm{sparse}^{\ast}$                  & 96.61          & 95.37          & 97.71          & 95.10
\end{tabular}
\caption{Results of Joint BERT~\citep{Chen2019} and different feature sets for DIET on the ATIS and SNIPS datasets. Reported numbers are accuracy for intents and micro-average F1 score for entities. The $\ast$ indicates that the data was annotated using the BILOU tagging schema. $\dagger$ implies that no mask loss was used. 
}\label{tab:results-snips-atis}
\end{table*}

In order to compare our results to the results presented in~\citet{Chen2019}, we use the same evaluation method as~\citet{Chen2019} and~\citet{goo2018slot}. They report the accuracy for intent classification and micro-averaged F1 score for entity recognition. Again, true positives, false positives, and false negatives for intent labels are obtained as in any other multi-class classification task. However, an entity only counts as a true positive if the prediction span exactly matches the gold span and their label match, a stricter definition than that of~\citet{vanzo2019hermit}. All experiments on ATIS and SNIPS were run 5 times. We take the average over the results from those runs as final numbers.  

To understand how transferable the hyperparameters of DIET are, we took the best performing model configurations of DIET on the NLU-Benchmark dataset and evaluated them on ATIS and SNIPS.
The intent classification accuracy and named entity recognition F1 score on the ATIS and SNIPS dataset are listed in Table~\ref{tab:results-snips-atis}. 

Due to the stricter evaluation method we tagged our data using the BILOU tagging schema~\citep{ramshaw1995text}.
The use of the BILOU tagging schmea is indicated by the $\ast$ in Table~\ref{tab:results-snips-atis}.

Remarkably, using only sparse features and no pre-trained embeddings whatsoever, DIET achieves performance within 1-2\% of the Joint BERT model.
Using the hyperparameters from the best performing model on the NLU-Benchmark dataset, DIET achieves results competitive with Joint BERT on both ATIS and SNIPS.

\section{Conclusion}
\label{sec:conclustion}

We introduced DIET, a flexible architecture for intent and entity modeling. We studied its performance on multiple datasets, and showed that DIET advances the state of the art on the challenging NLU-Benchmark dataset. Furthermore we extensively study the effectiveness of using embeddings from various pre-training methods. We find that there is no single set of embeddings which is always best across different datasets, highlighting the importance of a modular architecture. Furthermore we show that word embeddings from distributional models like GloVe are competitive with embeddings from large-scale language models, and that in fact without using any pre-trained embeddings, DIET can still achieve competitive performance, outperforming state of the art on NLU-Benchmark. Finally, we also show that the best set of pre-trained embeddings for DIET on NLU-Benchmark outperforms fine-tuning BERT inside DIET and is six times faster to train.


\bibliography{diet}

\begin{thebibliography}{48}
\expandafter\ifx\csname natexlab\endcsname\relax\def\natexlab#1{#1}\fi

\bibitem[{Abadi et~al.(2016)Abadi, Agarwal, Barham, Brevdo, Chen, Citro,
  Corrado, Davis, Dean, Devin, Ghemawat, Goodfellow, Harp, Irving, Isard, Jia,
  J{\'{o}}zefowicz, Kaiser, Kudlur, Levenberg, Man{\'{e}}, Monga, Moore,
  Murray, Olah, Schuster, Shlens, Steiner, Sutskever, Talwar, Tucker,
  Vanhoucke, Vasudevan, Vi{\'{e}}gas, Vinyals, Warden, Wattenberg, Wicke, Yu,
  and Zheng}]{abadi2015tensorflow}
Mart{\'{\i}}n Abadi, Ashish Agarwal, Paul Barham, Eugene Brevdo, Zhifeng Chen,
  Craig Citro, Gregory~S. Corrado, Andy Davis, Jeffrey Dean, Matthieu Devin,
  Sanjay Ghemawat, Ian~J. Goodfellow, Andrew Harp, Geoffrey Irving, Michael
  Isard, Yangqing Jia, Rafal J{\'{o}}zefowicz, Lukasz Kaiser, Manjunath Kudlur,
  Josh Levenberg, Dan Man{\'{e}}, Rajat Monga, Sherry Moore, Derek~Gordon
  Murray, Chris Olah, Mike Schuster, Jonathon Shlens, Benoit Steiner, Ilya
  Sutskever, Kunal Talwar, Paul~A. Tucker, Vincent Vanhoucke, Vijay Vasudevan,
  Fernanda~B. Vi{\'{e}}gas, Oriol Vinyals, Pete Warden, Martin Wattenberg,
  Martin Wicke, Yuan Yu, and Xiaoqiang Zheng. 2016.
\newblock \href {http://arxiv.org/abs/1603.04467} {Tensorflow: Large-scale
  machine learning on heterogeneous distributed systems}.
\newblock \emph{CoRR}, abs/1603.04467.

\bibitem[{Adhikari et~al.(2019)Adhikari, Ram, Tang, and Lin}]{docbert}
Ashutosh Adhikari, Achyudh Ram, Raphael Tang, and Jimmy Lin. 2019.
\newblock \href {http://arxiv.org/abs/1904.08398} {Docbert: {BERT} for document
  classification}.
\newblock \emph{CoRR}, abs/1904.08398.

\bibitem[{Baker et~al.(1998)Baker, Fillmore, and Lowe}]{baker97}
Collin~F. Baker, Charles~J. Fillmore, and John~B. Lowe. 1998.
\newblock \href {https://doi.org/10.3115/980451.980860} {The berkeley framenet
  project}.
\newblock In \emph{Proceedings of the 17th International Conference on
  Computational Linguistics - Volume 1}, COLING '98, pages 86--90, Stroudsburg,
  PA, USA. Association for Computational Linguistics.

\bibitem[{Casanueva et~al.(2020)Casanueva, Temčinas, Gerz, Henderson, and
  Vulić}]{casanueva2020efficient}
Iñigo Casanueva, Tadas Temčinas, Daniela Gerz, Matthew Henderson, and Ivan
  Vulić. 2020.
\newblock \href {http://arxiv.org/abs/2003.04807} {Efficient intent detection
  with dual sentence encoders}.

\bibitem[{Chen et~al.(2019)Chen, Zhuo, and Wang}]{Chen2019}
Qian Chen, Zhu Zhuo, and Wen Wang. 2019.
\newblock \href {http://arxiv.org/abs/1902.10909} {{BERT} for joint intent
  classification and slot filling}.
\newblock \emph{CoRR}, abs/1902.10909.

\bibitem[{Coucke et~al.(2018)Coucke, Saade, Ball, Bluche, Caulier, Leroy,
  Doumouro, Gisselbrecht, Caltagirone, Lavril, Primet, and
  Dureau}]{coucke2018snips}
Alice Coucke, Alaa Saade, Adrien Ball, Th{\'{e}}odore Bluche, Alexandre
  Caulier, David Leroy, Cl{\'{e}}ment Doumouro, Thibault Gisselbrecht,
  Francesco Caltagirone, Thibaut Lavril, Ma{\"{e}}l Primet, and Joseph Dureau.
  2018.
\newblock \href {http://arxiv.org/abs/1805.10190} {Snips voice platform: an
  embedded spoken language understanding system for private-by-design voice
  interfaces}.
\newblock \emph{CoRR}, abs/1805.10190.

\bibitem[{Devlin et~al.(2018)Devlin, Chang, Lee, and
  Toutanova}]{devlin2018bert}
Jacob Devlin, Ming-Wei Chang, Kenton Lee, and Kristina Toutanova. 2018.
\newblock Bert: Pre-training of deep bidirectional transformers for language
  understanding.
\newblock \emph{arXiv preprint arXiv:1810.04805}.

\bibitem[{Goo et~al.(2018)Goo, Gao, Hsu, Huo, Chen, Hsu, and
  Chen}]{goo2018slot}
Chih-Wen Goo, Guang Gao, Yun-Kai Hsu, Chih-Li Huo, Tsung-Chieh Chen, Keng-Wei
  Hsu, and Yun-Nung Chen. 2018.
\newblock \href {https://doi.org/10.18653/v1/N18-2118} {Slot-gated modeling for
  joint slot filling and intent prediction}.
\newblock In \emph{Proceedings of the 2018 Conference of the North {A}merican
  Chapter of the Association for Computational Linguistics: Human Language
  Technologies, Volume 2 (Short Papers)}, pages 753--757, New Orleans,
  Louisiana. Association for Computational Linguistics.

\bibitem[{Haihong et~al.(2019)Haihong, Niu, Chen, and Song}]{haihong2019anovel}
Ee~Haihong, Peiqing Niu, Zhongfu Chen, and Meina Song. 2019.
\newblock \href {http://arxiv.org/abs/1907.00390} {A novel bi-directional
  interrelated model for joint intent detection and slot filling}.
\newblock \emph{CoRR}, abs/1907.00390.

\bibitem[{Hemphill et~al.(1990)Hemphill, Godfrey, and
  Doddington}]{hemphill1990atis}
Charles~T. Hemphill, John~J. Godfrey, and George~R. Doddington. 1990.
\newblock \href {https://www.aclweb.org/anthology/H90-1021} {The {ATIS} spoken
  language systems pilot corpus}.
\newblock In \emph{Speech and Natural Language: Proceedings of a Workshop Held
  at Hidden Valley, {P}ennsylvania, June 24-27,1990}.

\bibitem[{Henderson et~al.(2019{\natexlab{a}})Henderson, Budzianowski,
  Casanueva, Coope, Gerz, Kumar, Mrksic, Spithourakis, Su, Vulic, and
  Wen}]{redditdata}
Matthew Henderson, Pawel Budzianowski, I{\~{n}}igo Casanueva, Sam Coope,
  Daniela Gerz, Girish Kumar, Nikola Mrksic, Georgios Spithourakis, Pei{-}Hao
  Su, Ivan Vulic, and Tsung{-}Hsien Wen. 2019{\natexlab{a}}.
\newblock \href {http://arxiv.org/abs/1904.06472} {A repository of
  conversational datasets}.
\newblock \emph{CoRR}, abs/1904.06472.

\bibitem[{Henderson et~al.(2019{\natexlab{b}})Henderson, Casanueva,
  Mrk{\v{s}}i{\'c}, Su, Vuli{\'c} et~al.}]{henderson2019convert}
Matthew Henderson, I{\~n}igo Casanueva, Nikola Mrk{\v{s}}i{\'c}, Pei-Hao Su,
  Ivan Vuli{\'c}, et~al. 2019{\natexlab{b}}.
\newblock Convert: Efficient and accurate conversational representations from
  transformers.
\newblock \emph{arXiv preprint arXiv:1911.03688}.

\bibitem[{Henderson et~al.(2019{\natexlab{c}})Henderson, Vuli{\'c}, Gerz,
  Casanueva, Budzianowski, Coope, Spithourakis, Wen, Mrk{\v{s}}i{\'c}, and
  Su}]{henderson2019training}
Matthew Henderson, Ivan Vuli{\'c}, Daniela Gerz, I{\~n}igo Casanueva, Pawe{\l}
  Budzianowski, Sam Coope, Georgios Spithourakis, Tsung-Hsien Wen, Nikola
  Mrk{\v{s}}i{\'c}, and Pei-Hao Su. 2019{\natexlab{c}}.
\newblock Training neural response selection for task-oriented dialogue
  systems.
\newblock \emph{arXiv preprint arXiv:1906.01543}.

\bibitem[{Howard and Ruder(2018)}]{howardruder}
Jeremy Howard and Sebastian Ruder. 2018.
\newblock \href {http://arxiv.org/abs/1801.06146} {Fine-tuned language models
  for text classification}.
\newblock \emph{CoRR}, abs/1801.06146.

\bibitem[{Japkowicz and Stephen(2002)}]{japkowicz2002class}
Nathalie Japkowicz and Shaju Stephen. 2002.
\newblock The class imbalance problem: A systematic study.
\newblock \emph{Intelligent data analysis}, 6(5):429--449.

\bibitem[{Johnson(2018)}]{khari}
Khari Johnson. 2018.
\newblock Facebook messenger passes 300,000 bots.
\newblock
  \href{https://venturebeat.com/2018/05/01/facebook-messenger-passes-300000-bots/}{venturebeat.com}
  {[Online; posted 1-May-2018]}.

\bibitem[{Kingma and Ba(2014)}]{Adam2014}
Diederik~P. Kingma and Jimmy Ba. 2014.
\newblock \href {http://arxiv.org/abs/1412.6980} {Adam: A method for stochastic
  optimization}.
\newblock Cite arxiv:1412.6980Comment: Published as a conference paper at the
  3rd International Conference for Learning Representations, San Diego, 2015.

\bibitem[{Klein and Nabi(2019)}]{commonsensebert}
Tassilo Klein and Moin Nabi. 2019.
\newblock \href {http://arxiv.org/abs/1905.13497} {Attention is (not) all you
  need for commonsense reasoning}.
\newblock \emph{CoRR}, abs/1905.13497.

\bibitem[{Lafferty et~al.(2001)Lafferty, McCallum, and
  Pereira}]{lafferty2001conditional}
John Lafferty, Andrew McCallum, and Fernando~CN Pereira. 2001.
\newblock Conditional random fields: Probabilistic models for segmenting and
  labeling sequence data.

\bibitem[{Lample et~al.(2016)Lample, Ballesteros, Subramanian, Kawakami, and
  Dyer}]{crfloss}
Guillaume Lample, Miguel Ballesteros, Sandeep Subramanian, Kazuya Kawakami, and
  Chris Dyer. 2016.
\newblock \href {http://arxiv.org/abs/1603.01360} {Neural architectures for
  named entity recognition}.
\newblock \emph{CoRR}, abs/1603.01360.

\bibitem[{Lee and Hsiang(2019)}]{patentbert}
Jieh{-}Sheng Lee and Jieh Hsiang. 2019.
\newblock \href {http://arxiv.org/abs/1906.02124} {Patentbert: Patent
  classification with fine-tuning a pre-trained {BERT} model}.
\newblock \emph{CoRR}, abs/1906.02124.

\bibitem[{Liu and Lane(2016)}]{liu2016attention}
Bing Liu and Ian Lane. 2016.
\newblock \href {http://arxiv.org/abs/1609.01454} {Attention-based recurrent
  neural network models for joint intent detection and slot filling}.
\newblock \emph{CoRR}, abs/1609.01454.

\bibitem[{Liu et~al.(2019{\natexlab{a}})Liu, He, Chen, and Gao}]{mtdnn2019}
Xiaodong Liu, Pengcheng He, Weizhu Chen, and Jianfeng Gao. 2019{\natexlab{a}}.
\newblock \href {http://arxiv.org/abs/1904.09482} {Improving multi-task deep
  neural networks via knowledge distillation for natural language
  understanding}.
\newblock \emph{CoRR}, abs/1904.09482.

\bibitem[{Liu et~al.(2019{\natexlab{b}})Liu, Eshghi, Swietojanski, and
  Rieser}]{liu2019benchmarking}
Xingkun Liu, Arash Eshghi, Pawel Swietojanski, and Verena Rieser.
  2019{\natexlab{b}}.
\newblock \href {http://arxiv.org/abs/1903.05566} {Benchmarking natural
  language understanding services for building conversational agents}.
\newblock \emph{CoRR}, abs/1903.05566.

\bibitem[{Mehri et~al.(2019)Mehri, Srinivasan, and
  Eskenazi}]{mehri2019structured}
Shikib Mehri, Tejas Srinivasan, and Maxine Eskenazi. 2019.
\newblock Structured fusion networks for dialog.
\newblock \emph{arXiv preprint arXiv:1907.10016}.

\bibitem[{Pennington et~al.(2014)Pennington, Socher, and
  Manning}]{pennington2014glove}
Jeffrey Pennington, Richard Socher, and Christopher Manning. 2014.
\newblock Glove: Global vectors for word representation.
\newblock In \emph{Proceedings of the 2014 conference on empirical methods in
  natural language processing (EMNLP)}, pages 1532--1543.

\bibitem[{Peters et~al.(2018)Peters, Neumann, Iyyer, Gardner, Clark, Lee, and
  Zettlemoyer}]{elmo}
Matthew~E. Peters, Mark Neumann, Mohit Iyyer, Matt Gardner, Christopher Clark,
  Kenton Lee, and Luke Zettlemoyer. 2018.
\newblock \href {http://arxiv.org/abs/1802.05365} {Deep contextualized word
  representations}.
\newblock \emph{CoRR}, abs/1802.05365.

\bibitem[{Peters et~al.(2019)Peters, Ruder, and Smith}]{ruder2019tuning}
Matthew~E. Peters, Sebastian Ruder, and Noah~A. Smith. 2019.
\newblock \href {http://arxiv.org/abs/1903.05987} {To tune or not to tune?
  adapting pretrained representations to diverse tasks}.
\newblock \emph{CoRR}, abs/1903.05987.

\bibitem[{Radford(2018)}]{gpt}
Alec Radford. 2018.
\newblock Improving language understanding by generative pre-training.

\bibitem[{Ramshaw and Marcus(1995)}]{ramshaw1995text}
Lance~A. Ramshaw and Mitchell~P. Marcus. 1995.
\newblock \href {http://arxiv.org/abs/cmp-lg/9505040} {Text chunking using
  transformation-based learning}.
\newblock \emph{CoRR}, cmp-lg/9505040.

\bibitem[{Shaw et~al.(2018)Shaw, Uszkoreit, and Vaswani}]{shaw2018self}
Peter Shaw, Jakob Uszkoreit, and Ashish Vaswani. 2018.
\newblock Self-attention with relative position representations.
\newblock \emph{arXiv preprint arXiv:1803.02155}.

\bibitem[{Smith et~al.(2017)Smith, Kindermans, Ying, and Le}]{smith2017don}
Samuel~L Smith, Pieter-Jan Kindermans, Chris Ying, and Quoc~V Le. 2017.
\newblock Don't decay the learning rate, increase the batch size.
\newblock \emph{arXiv preprint arXiv:1711.00489}.

\bibitem[{Strubell et~al.(2019)Strubell, Ganesh, and
  McCallum}]{strubell2019energy}
Emma Strubell, Ananya Ganesh, and Andrew McCallum. 2019.
\newblock Energy and policy considerations for deep learning in nlp.
\newblock \emph{arXiv preprint arXiv:1906.02243}.

\bibitem[{Sun et~al.(2019{\natexlab{a}})Sun, Myers, Vondrick, Murphy, and
  Schmid}]{videobert}
Chen Sun, Austin Myers, Carl Vondrick, Kevin Murphy, and Cordelia Schmid.
  2019{\natexlab{a}}.
\newblock \href {http://arxiv.org/abs/1904.01766} {Videobert: {A} joint model
  for video and language representation learning}.
\newblock \emph{CoRR}, abs/1904.01766.

\bibitem[{Sun et~al.(2019{\natexlab{b}})Sun, Qiu, Xu, and
  Huang}]{howtotunebert}
Chi Sun, Xipeng Qiu, Yige Xu, and Xuanjing Huang. 2019{\natexlab{b}}.
\newblock \href {http://arxiv.org/abs/1905.05583} {How to fine-tune {BERT} for
  text classification?}
\newblock \emph{CoRR}, abs/1905.05583.

\bibitem[{Taylor(1953)}]{taylor1953cloze}
Wilson~L Taylor. 1953.
\newblock “cloze procedure”: A new tool for measuring readability.
\newblock \emph{Journalism Bulletin}, 30(4):415--433.

\bibitem[{Vanzo et~al.(2019)Vanzo, Bastianelli, and Lemon}]{vanzo2019hermit}
Andrea Vanzo, Emanuele Bastianelli, and Oliver Lemon. 2019.
\newblock \href {https://doi.org/10.18653/v1/W19-5931} {Hierarchical multi-task
  natural language understanding for cross-domain conversational ai: Hermit
  nlu}.
\newblock pages 254--263.

\bibitem[{Varghese et~al.(2020)Varghese, Sarang, Yadav, Karotra, and
  Gandhi}]{jointvarghese}
Akson~Sam Varghese, Saleha Sarang, Vipul Yadav, Bharat Karotra, and Niketa
  Gandhi. 2020.
\newblock Bidirectional lstm joint model for intent classification and named
  entity recognition in natural language understanding.
\newblock In \emph{Intelligent Systems Design and Applications}, pages 58--68,
  Cham. Springer International Publishing.

\bibitem[{Vaswani et~al.(2017)Vaswani, Shazeer, Parmar, Uszkoreit, Jones,
  Gomez, Kaiser, and Polosukhin}]{vaswani2017attention}
Ashish Vaswani, Noam Shazeer, Niki Parmar, Jakob Uszkoreit, Llion Jones,
  Aidan~N Gomez, {\L}ukasz Kaiser, and Illia Polosukhin. 2017.
\newblock Attention is all you need.
\newblock In \emph{Advances in neural information processing systems}, pages
  5998--6008.

\bibitem[{Vlasov et~al.(2019)Vlasov, Mosig, and Nichol}]{vlasov2019dialogue}
Vladimir Vlasov, Johannes~EM Mosig, and Alan Nichol. 2019.
\newblock Dialogue transformers.
\newblock \emph{arXiv preprint arXiv:1910.00486}.

\bibitem[{Wang et~al.(2019)Wang, Pruksachatkun, Nangia, Singh, Michael, Hill,
  Levy, and Bowman}]{superglue2019}
Alex Wang, Yada Pruksachatkun, Nikita Nangia, Amanpreet Singh, Julian Michael,
  Felix Hill, Omer Levy, and Samuel~R. Bowman. 2019.
\newblock \href {http://arxiv.org/abs/1905.00537} {Superglue: {A} stickier
  benchmark for general-purpose language understanding systems}.
\newblock \emph{CoRR}, abs/1905.00537.

\bibitem[{Wang et~al.(2018)Wang, Singh, Michael, Hill, Levy, and Bowman}]{glue}
Alex Wang, Amanpreet Singh, Julian Michael, Felix Hill, Omer Levy, and
  Samuel~R. Bowman. 2018.
\newblock \href {http://arxiv.org/abs/1804.07461} {{GLUE:} {A} multi-task
  benchmark and analysis platform for natural language understanding}.
\newblock \emph{CoRR}, abs/1804.07461.

\bibitem[{Williams et~al.(2017)Williams, Asadi, and Zweig}]{williams2017hybrid}
Jason~D Williams, Kavosh Asadi, and Geoffrey Zweig. 2017.
\newblock Hybrid code networks: practical and efficient end-to-end dialog
  control with supervised and reinforcement learning.
\newblock \emph{arXiv preprint arXiv:1702.03274}.

\bibitem[{Williams and Young(2007)}]{williams2007partially}
Jason~D Williams and Steve Young. 2007.
\newblock Partially observable markov decision processes for spoken dialog
  systems.
\newblock \emph{Computer Speech \& Language}, 21(2):393--422.

\bibitem[{Wu et~al.(2017)Wu, Fisch, Chopra, Adams, Bordes, and
  Weston}]{wu2017starspace}
Ledell Wu, Adam Fisch, Sumit Chopra, Keith Adams, Antoine Bordes, and Jason
  Weston. 2017.
\newblock Starspace: Embed all the things!
\newblock \emph{arXiv preprint arXiv:1709.03856}.

\bibitem[{Yoshihashi et~al.(2018)Yoshihashi, Shao, Kawakami, You, Iida, and
  Naemura}]{class-reconstruct2019}
Ryota Yoshihashi, Wen Shao, Rei Kawakami, Shaodi You, Makoto Iida, and Takeshi
  Naemura. 2018.
\newblock \href {http://arxiv.org/abs/1812.04246}
  {Classification-reconstruction learning for open-set recognition}.
\newblock \emph{CoRR}, abs/1812.04246.

\bibitem[{Zhang and Wang(2016)}]{zhang2016ajoint}
Xiaodong Zhang and Houfeng Wang. 2016.
\newblock \href {http://dl.acm.org/citation.cfm?id=3060832.3061040} {A joint
  model of intent determination and slot filling for spoken language
  understanding}.
\newblock In \emph{Proceedings of the Twenty-Fifth International Joint
  Conference on Artificial Intelligence}, IJCAI'16, pages 2993--2999. AAAI
  Press.

\bibitem[{Zhang et~al.(2019)Zhang, Han, Liu, Jiang, Sun, and Liu}]{ernie2019}
Zhengyan Zhang, Xu~Han, Zhiyuan Liu, Xin Jiang, Maosong Sun, and Qun Liu. 2019.
\newblock \href {http://arxiv.org/abs/1905.07129} {{ERNIE:} enhanced language
  representation with informative entities}.
\newblock \emph{CoRR}, abs/1905.07129.

\end{thebibliography}
\bibliographystyle{acl_natbib}

\end{document}